\documentclass[11pt,a4paper]{article}
\usepackage{times,latexsym}
\usepackage{url}
\usepackage[T1]{fontenc}

  \usepackage[acceptedWithA]{tacl2021v1}

\usepackage{microtype}

\usepackage{inconsolata}

\usepackage{graphicx}
\usepackage{subcaption} %

\usepackage{pifont}%
\newcommand{\cmark}{\ding{51}}%
\newcommand{\xmark}{\ding{55}}%

\usepackage{booktabs} %
\usepackage{listings} %
\usepackage{tikz} %
\usetikzlibrary{patterns}
\usepackage{pgfplots} %
\pgfplotsset{compat=1.17} %
\definecolor{bblue}{HTML}{4F81BD}
\definecolor{rred}{HTML}{C0504D}
\definecolor{ggreen}{HTML}{9BBB59}
\definecolor{ppurple}{HTML}{9F4C7C}

\usepackage[]{tacl2021v1}

\usepackage{xspace,mfirstuc,tabulary}

\newif\iftaclinstructions
\taclinstructionsfalse %
\iftaclinstructions
\renewcommand{\confidential}{}
\renewcommand{\anonsubtext}{(No author info supplied here, for consistency with
TACL-submission anonymization requirements)}
\newcommand{\instr}
\fi

\iftaclpubformat %

\else

\fi

\title{Constructing Multilingual Visual-Text Datasets Revealing Visual Multilingual Ability of Vision Language Models}

\author{
  Jesse Atuhurra$^\spadesuit$ Iqra Ali$^\spadesuit$ Tatsuya Hiraoka$^\dagger$ Hidetaka Kamigaito$^\spadesuit$ \\ 
  \textbf{Tomoya Iwakura$^\dagger$ Taro Watanabe$^\spadesuit$}
\\
  $^\spadesuit$ Information Science Division, NAIST, Japan
  $^\dagger$Fujitsu Limited, Japan
  \\
 \texttt{ \{atuhurra.jesse.ag2, ali.iqra.ai6, kamigaito.h, taro\}@naist.ac.jp } 
 \\
 \texttt{ \{hiraoka.tatsuya, iwakura.tomoya\}@fujitsu.com }
  \\
}

\date{}

\begin{document}
\maketitle

\begin{abstract}
Large language models (LLMs) have increased interest in vision language models (VLMs), which process image-text pairs as input. 
Studies investigating the visual understanding ability of VLMs have been proposed, but such studies are still preliminary because existing datasets do not permit a comprehensive evaluation of the fine-grained visual linguistic abilities of VLMs across multiple languages.
To further explore the strengths of VLMs, such as GPT-4V~\cite{openai2023GPT4}, we developed new datasets for the systematic and qualitative analysis of VLMs. Our contribution is four-fold: 1) we introduced nine vision-and-language (VL) tasks (including object recognition, image-text matching, and more) and constructed multilingual visual-text datasets in four languages: English, Japanese, Swahili, and Urdu through utilizing templates containing \textit{questions} and prompting GPT4-V to generate the \textit{answers} and the \textit{rationales}, 2) introduced a new VL task named \textit{unrelatedness}, 3) introduced rationales to enable human understanding of the VLM reasoning process, and 4) employed human evaluation to measure the suitability of proposed datasets for VL tasks. We show that VLMs can be fine-tuned on our datasets. Our work is the first to conduct such analyses in Swahili and Urdu. Also, it introduces \textit{rationales} in VL analysis, which played a vital role in the evaluation.
\end{abstract}

\section{Introduction}
\label{sec:Introduction}
Large language models (LLMs), a particular class of foundation models (FMs), have taken the natural language processing (NLP) community by storm. The rise in popularity of LLMs is credited to the development of the Transformer~\cite{10.5555/3295222.3295349} architecture in 2017 and ELMo \cite{peters-etal-2018-deep} in 2018. Since then, a plethora of LLMs have been developed.  Despite their ability to solve several NLP tasks, LLMs were not ``mainstream'' until the release of ChatGPT in November 2022. The arrival of ChatGPT meant that many non-tech-savvy users could leverage this tool to accomplish several day-to-day tasks. However, it soon became clear that the ``text-only'' input of ChatGPT was a major limitation \cite{qin-etal-2023-chatgpt}, \cite{yin2023survey}. 
\begin{figure}[!t]
\centering
\includegraphics[width=0.95\linewidth]{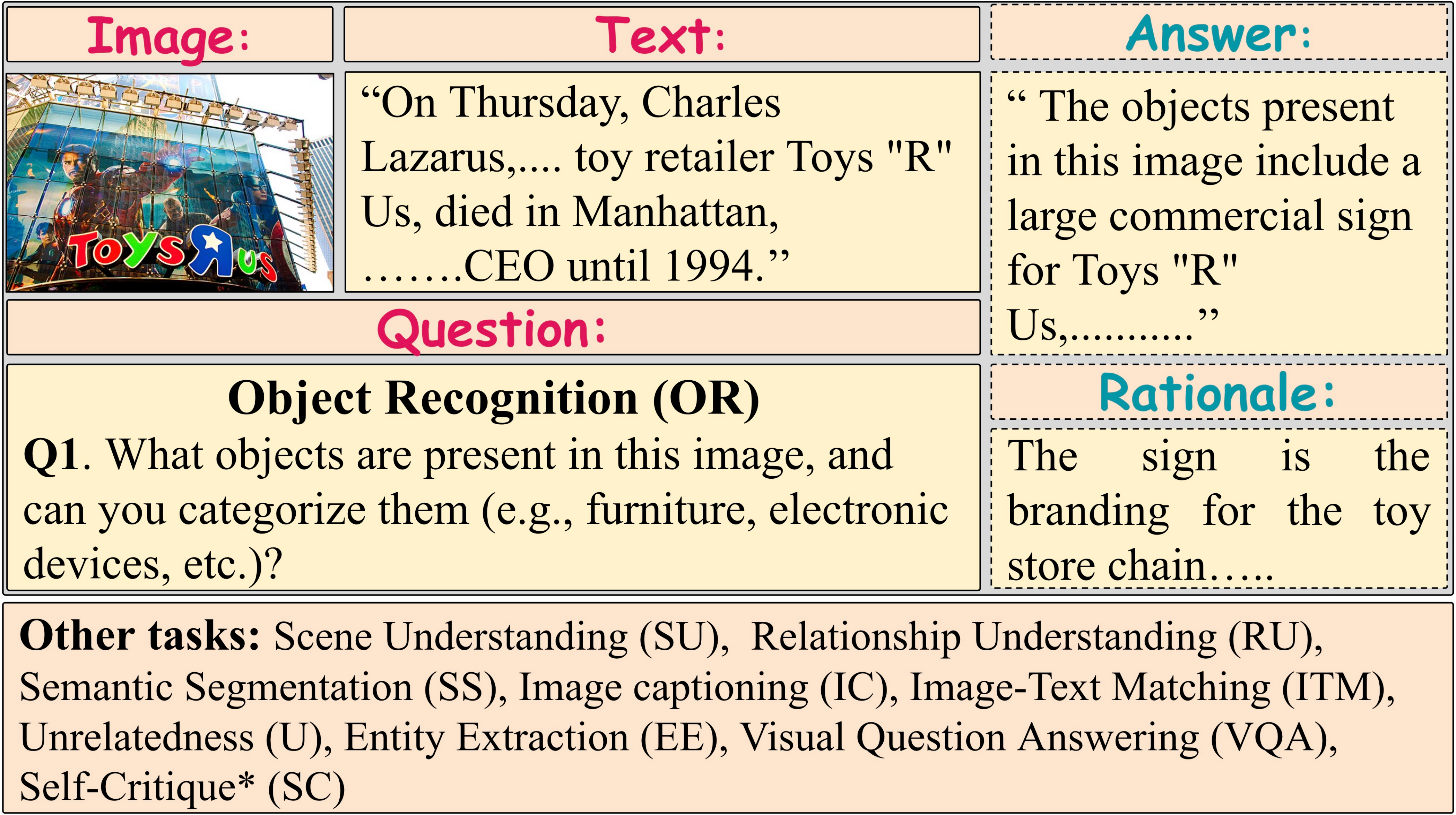} 
\caption{To study the visual and multilingual abilities of VLMs (i.e., GPT-4V), we introduced nine tasks. The input to GPT-4V includes \textbf{image, text, questions} while it outputs an \textbf{answer, rationale} pair.
We repeated this process in \textbf{English (En), Japanese (Jp), Urdu (Ur), Swahili (Sw)}, and constructed datasets in these four languages. (An expanded version in Figure \ref{fig:IntroFigureExpanded} in Appendix~\ref{Appendix:IntroFigureExpanded}).} 
\label{fig:IntroFigure}
\end{figure}
NLP researchers started creating new architectures that would make it possible for FMs to process other types of \textit{modalities}\footnote{Such FMs are called large multimodal models (LMMs).}. Specifically, the need for FMs that accept both image and text input led to the development of vision language models (VLMs). VLMs are a class of FMs or LMMs\footnote{It is worth noting that the jargon ``LLM, FM, VLM'' are often used interchangeably in NLP literature.}  which mainly comprise of two \textit{backbones}, one for image representation and the other for text representation. 

Benchmarks consisting of many tasks have been developed to unravel VLMs' visual and linguistic abilities. Some benchmarks are text-only, for example, MMLU~\cite{hendrycks2021measuring} and AGIEval~\cite{zhong2023agieval}, while others are text+image benchmarks. These include MME~\cite{fu2023mme}, MMBench~\cite{liu2023mmbench}, and SEED-Bench~\cite{li2023seedbench}. NLP researchers have previously evaluated VLMs against these benchmarks.   

While several studies have investigated the VL ability of VLMs, studies regarding multilingual ability are still preliminary for two reasons. \emph{First.} Available datasets/benchmarks i) focus mainly on ``high resource languages'' like English with little exploration of ``under-served languages'', ii) are described by short captions in the image-text pairs, iii) and suffer from limited diversity of image samples (Section~\ref{subsec: Benchmarks for VLMs}), which inhibits our ability to thoroughly investigate the true abilities of VLMs to decipher the visual information contained in images across languages and domains. \emph{Second.} VLMs too, especially the open-source VLMs, support only the high resource languages (see Section~\ref{subsec:State-of-the-art VLMs}). %

To tackle these challenges, we introduce i)  the first VL datasets for both Swahili, a Sub-Saharan language, and Urdu, a South Asian language, in addition to English and Japanese, ii) datasets which have \textit{rich captions}\footnote{The image-text pairs in this study are characterized by ``rich text'' containing several paragraphs, rather than brief phrases as in several existing datasets.}, iii) datasets covering ten broad image-text categories. 

In this study, our goal is to uncover the fine-grained VL abilities of VLMs. We divide this goal into sub-goals: 1) introduce VL tasks to measure VLM strengths, 2) reduce the language gap in available datasets by developing new \textit{multilingual} datasets (see Figure \ref{fig:IntroFigure}) for the systematic and qualitative analysis of VLMs. Throughout this process, we revealed the performance gap across four languages, and 3) show that the proposed dataset is effective for fine-tuning VLMs. Our work is the first to construct such datasets and conduct VL analyses in Swahili and Urdu.

The main VLM in this study is GPT-4V because, based on many benchmarks and analysis by ~\newcite{akter2023indepth}, GPT-4V is the most powerful VLM as of this writing. We leveraged its multimodal abilities to create all datasets in our study. 

These are the main contributions of our work:
\begin{enumerate}
    \item We introduce nine vision-and-language tasks and construct datasets containing image-text pairs in English, Japanese, Swahili, and Urdu (Section \ref{sec:Datasets}). %
    \item We introduce a new task, dubbed \textit{unrelatedness}, which measures VLM ability to identify parts of \textit{text} that are ``not related'' to the image (Section \ref{subsec:VisionAndLanguageTasks}).  
    \item We introduce \textit{rationales} in our datasets (See Figure \ref{fig:Input2GPT4}). The rationales facilitate human evaluation in this study. %
    \item We leverage human evaluation by recruiting native speakers (Sections \ref{subsec:HumanEvaluation} and \ref{sec:Results}) for each language to measure the suitability of our datasets for VL tasks. %
\end{enumerate}
\section{Related Work}
\label{sec:RelatedWork}
\subsection{State-of-the-art VLMs} 
\label{subsec:State-of-the-art VLMs}
We can categorize the most potent VLMs into closed-source and open-source VLMs. Major closed-source VLMs are: GPT-4V, Gemini~\cite{gemini_2023}, Qwen-VL~\cite{Qwen-VL}, and Kosmos-2 \cite{peng2023kosmos2},  while prominent open-source VLMs include: LLaVa~\cite{liu2024llava16, liu2023improvedllava, liu2023llava},  mPLUG-Owl \cite{ye2023mplugowl2}, \cite{ye2023mplugowl}, CogVLM \cite{wang2023cogvlm}, BLIP-2~\cite{li2023blip2}, MiniGPT-4~\cite{zhu2023miniGPT4}, InstructBLIP~\cite{dai2023instructblip}, Shikra \cite{chen2023shikra}, OpenFlamingo \cite{awadalla2023openflamingo, anas_awadalla_2023_7733589, Alayrac2022FlamingoAV} and many more. One major limitation stands out in these VLMs; they do not support \textit{multiple} languages,  except GPT-4V\footnote{Unofficial list of languages in GPT-4V \url{https://www.mlyearning.org/languages-supported-by-chatgpt/}.} and Gemini\footnote{Official list of languages supported by Gemini \url{https://ai.google.dev/available_regions}.}. Regarding our study, GPT-4V supports all four languages, namely English, Japanese, Swahili, and Urdu, yet Gemini supports all the languages except Urdu. 
\subsection{In-context Learning} In-context learning or ICL \cite{NEURIPS2020_1457c0d6} is a revolutionary approach in NLP that empowers LLMs to adapt and learn new tasks without extensive training. Instead of requiring large datasets and complex fine-tuning, ICL allows us to learn new skills on the fly by simply providing instructions and examples within the task's context. We leveraged ICL in this study by creating prompts for all the VL tasks. 
\subsection{Relevance of Rationales in LMMs} Rationales serve as a gateway towards understanding the thought processes of LMMs. Rationales explain how LMMs arrived at the final answer \cite{ling-etal-2017-program, hsieh-etal-2023-distilling}. Moreover, rationales assist in creating better prompting methods. In this study, we prompted the VLM to generate the rationale for each answer on top of the answer itself, which was valuable at the human evaluation stage.
\begin{table*}[!t]
\footnotesize
\centering
\resizebox{\textwidth}{!}{
\begin{tabular}{lllllrr}
\toprule
\textbf{Dataset} & \textbf{Task}  & \textbf{Multilingual} & \textbf{Language(s)} & \textbf{Rationales} & \textbf{\#Images} & \textbf{\#Questions} \\
\midrule
VQAv2~\cite{goyal2017making}        & VQA  & \xmark & En  & \xmark & 265K & 1.1M \\ %
OK-VQA~\cite{okvqa}       & VQA  & \xmark & En  & \xmark & 14K & 14K  \\ %
OCR-VQA~\cite{mishraICDAR19}   & VQA  & \xmark & En  & \xmark & 207K & 1M \\ %
GQA~\cite{hudson2019gqa}   & VQA  & \xmark & En  & \xmark & 113K & 22M \\ %
Visual Genome \cite{Krishna2016VisualGC} & VQA  & \xmark & En  & \xmark & 108K & 1.7M \\ %
VizWizQA \cite{gurari2018vizwiz}  & VQA & \xmark & En & \xmark & * & 31.1K \\ %
TextVQA \cite{singh2019towards}   & VQA & \xmark & En & \xmark & 28K & 45.3K \\ %
\midrule
LAION 5B~\cite{schuhmann2022laion5b}     & IC  & \cmark & En, Zh,..  & \xmark &  5.85B & * \\ %
LAION-COCO & IC & \xmark & En  & \xmark & 600M & *\\ %
Visual Genome \cite{Krishna2016VisualGC}  & IC & \xmark & En  & \xmark & 108KM & 1.77M \\
MSCOCO~\cite{lin2015microsoft}  & IC & \xmark & En  & \xmark & 328K & *\\
Flickr30k~\cite{plummer2016flickr30k} & IC & \xmark & En  & \xmark & 30K & * \\
Crossmodal-3600~\cite{thapliyal-etal-2022-crossmodal}  & IC & \cmark (36) & En, Jp, Sw, ..  & \xmark & 3.6K & *\\
\midrule
RefClef \cite{kazemzadeh-etal-2014-referitgame} & REG & \xmark & En & \xmark & 19.9K & * \\
RefCOCO \cite{kazemzadeh-etal-2014-referitgame}  & REG & \xmark & En & \xmark & 19.9K & *\\
RefCOCO+ \cite{kazemzadeh-etal-2014-referitgame}  & REG & \xmark & En & \xmark & 19.9K & *\\
RefCOCOg \cite{mao2016generation}  & REG & \xmark & En & \xmark & 25.7K & *\\
\midrule
MMMU \cite{yue2023mmmu} & VQA &  \xmark & En & \xmark & 11K & 11.5K \\ %
MME$+$ \cite{fu2023mme} & OR, OCR, ..  &  \xmark & En & \xmark & 1K & 2K\\ %
MMBench~\cite{liu2023mmbench}& OCR, .. & \cmark & En, Zh & \xmark & 2.9K & 2.9K \\ %
SEED-Bench~\cite{li2023seedbench} & OCR,.. &  \xmark & En & \xmark & 19K & 19K \\ %
MathVista \cite{lu2024mathvista}   & Maths &  \xmark & En & \xmark & 6.1K & 6.1K\\ %
MM-Vet \cite{yu2023mmvet} & OCR,OR,.. &  \xmark & En & \xmark & 200 & 218 \\ %
Q-Bench \cite{wu2024qbench}  & VQA & \xmark & En & \xmark & 3.4K & 2.9K \\ %
\midrule
\textbf{Ours} &  OR,IC,VQA,.. & \cmark (4) & En, Jp, Sw, Ur  & \cmark & 721 & 7.2K \\
\bottomrule
\end{tabular} }
\caption{\label{Table:VLMBenchmarks} A comparison between our dataset and existing datasets for visual question answering (VQA), image captioning (IC), and reference expression generation (REG). Datasets in the bottom row cover several tasks. MME tasks are object recognition (OR), optical character recognition (OCR), commonsense reasoning, numerical calculation, text translation, and code reasoning. MMBench comprises 20 sub-tasks. SEED-Bench includes 12 sub-tasks. MathVista includes maths questions from 28 datasets. MM-Vet evaluated 6 VL abilities. Q-Bench measures the low-level perception, description, and assessment abilities of VLMs. More MMBench, SEED-Bench, and MM-Vet details are in Appendix \ref{Appendix:benchmarkdetails}.
In this table \cmark~means ``availabe'' while \xmark~means ``unavailable''. Yet * means the original paper did not explicitly mention the number of images or questions. Compared to all these datasets, only our dataset introduces rationales.}
\end{table*}
\subsection{Datasets and Benchmarks for VLMs}
\label{subsec: Benchmarks for VLMs}
These are efforts to develop VL datasets specific to image captioning (IC), reference expression generation (REG), reference expression comprehension (REC), visual question answering (VQA), and more (See Table \ref{Table:VLMBenchmarks}).  \\
\textbf{Image Captioning.} Popular datasets are: MSCOCO~\cite{lin2015microsoft}, Flickr30k~\cite{plummer2016flickr30k}, Conceptual 12M~\cite{changpinyo2021cc12m}, Conceptual Captions~\cite{sharma2018conceptual}, all of which are in English. In response to the growing need for image captions in other languages, the Multi30K~\cite{elliott-etal-2016-multi30k} dataset extended Flickr30k to German, English, French, and Czech. Then, \newcite{ilahi2021efficient} extended a portion of Flickr8k~\cite{2fc2fb1303244a3b896f6f38eeeae76b} to Urdu. MSCOCO was extended to Japanese in the STAIR Captions~\cite{Yoshikawa2017} and  YJ Captions~\cite{P16-1168} datasets. AI Challenger~\cite{Wu_2019} is available in Chinese. Crossmodal-3600~\cite{thapliyal-etal-2022-crossmodal} is available in 36 languages, including English, Japanese, and Swahili. However, significant datasets are in high-resource languages, leaving room for further exploration in low-resource languages. The Swahili data in Crossmodal-3600 contains only 100 images and two captions per image. The dataset by \newcite{ilahi2021efficient} contains 700 images, limited only to images containing ``people''. \textit{In both datasets, Crossmodal-3600 and \cite{ilahi2021efficient}, which include Swahili and Urdu, respectively, the captions do not give thorough details about complex scenes in the image}. Another problem is the lack of a balanced representation of image categories. \\
\textbf{Reference Expression Generation.} Most notable REG\footnote{The REG task aims to provide a natural language description for a \textit{specific object} within the image, which is contrary to the IC task, which aims to condition the generation of natural language on all the visual features contained in the image. On the other hand, REC aims to \textit{identify} the referred object in the image when the input is an expression in natural language.} datasets are RefCOCO \cite{kazemzadeh-etal-2014-referitgame}, RefCOCO+ \cite{kazemzadeh-etal-2014-referitgame}, and RefCOCOg \cite{mao2016generation}, all of which suffer from the limitation that the description for the target object in the image is too short. \\
\textbf{Visual Question Answering.} In the same vein as REG, datasets developed for VQA contain limited descriptive information \cite{goyal2017making, okvqa, mishraICDAR19, hudson2019gqa, Krishna2016VisualGC} about the objects contained in the image. Additionally, the datasets are predominantly \textit{English}, leaving room for developing datasets in underserved languages, such as Swahili and Urdu. \\
\textbf{Multimodal Evaluation Benchmarks.} Specific to the evaluation of LLMs and LMMs, many benchmarks have been developed to enable transparent evaluation of VLMs' visual/multimodal understanding abilities. The benchmarks are: text-only, such as ARC \cite{clark2018think}, MMLU~\cite{hendrycks2021measuring}, BIG-Bench Hard \cite{suzgun2022challenging}, DROP \cite{dua-etal-2019-drop}, HellaSwag \cite{zellers-etal-2019-hellaswag}, GSM8K \cite{cobbe2021training}, MATH \cite{hendrycks2021measuring}, HumanEval \cite{chen2021evaluating}, and AGIEval~\cite{zhong2023agieval}: text-and-image, such as MMMU \cite{yue2023mmmu}, TextVQA \cite{singh2019towards}, MathVista \cite{lu2024mathvista}, MME~\cite{fu2023mme}, MMBench~\cite{liu2023mmbench}, SEED-Bench~\cite{li2023seedbench}; video, such as Perception Test \cite{patraucean2023perception}. We dissociate with these benchmarks and construct multilingual multimodal datasets containing image-text pairs mainly from Wikinews articles and occasionally from Wikipedia. (More details about this decision are described in Section~\ref{sec:Data source}.) As shown above, IC, REG, and VQA datasets did not introduce information, which we called ``rationales'', about the \textit{caption, reference expression, or answer} generated/provided by the model or human. These drawbacks motivated our efforts to construct VL datasets in this study (Refer to Table~\ref{Table:VLMBenchmarks} for comparing prior datasets and our proposed dataset). \\
\subsection{Other Concurrent Works.} 
\label{subsec:otherworks}
Recent attempts to investigate the visual and linguistic prowess of GPT-4V include \newcite{wu2023GPT4vis} who leveraged the textual descriptions generated by GPT-4V about an image, video, or point cloud to improve zero-shot visual recognition with CLIP; and \newcite{wu2023early} who studied several abilities of GPT-4V, such as visual understanding, language understanding, and visual puzzle solving. All these works rely on existing benchmarks for their evaluation. The benchmarks are mainly limited to English or Chinese, without considering other languages like Swahili or Urdu.
\begin{figure*}[!t]
\centering
\includegraphics[width=0.82\linewidth]{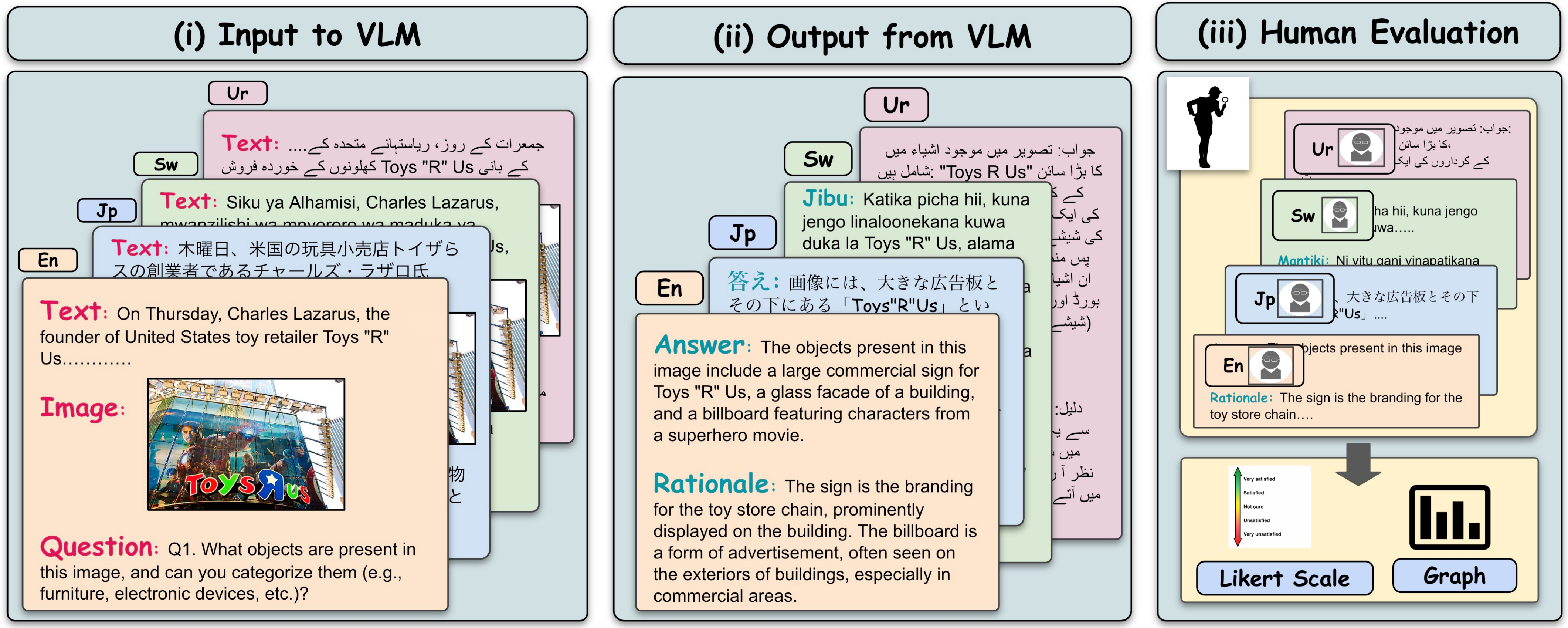} 
\caption{During dataset construction, the VLM input is a prompt which consists of \textbf{text, image, question}. The output from the VLM is the \textbf{answer, rationale} pair. Humans, specifically native speakers of \textbf{En, Jp, Sw, Ur} respectively, evaluate the quality of the answer taking into consideration the rationale generated by the VLM. }
\label{fig:Input2GPT4}
\end{figure*}
\section{Datasets}
\label{sec:Datasets}
To construct the datasets, we leveraged ICL and designed prompts (see Figure \ref{fig:Input2GPT4}) relevant to perform nine tasks.\footnote{This is made possible by multiple interpretations of pixel information inside one image. Such interpretations permit for a full capture of the VLM's fine-grained abilities\cite{liu2023mmbench}. Notably, current VLMs are powerful enough that we can investigate several VL abilities using the same image.} The tasks include Object Recognition (OR), Scene Understanding (SU), Relationship Understanding (RU), Semantic Segmentation (SS), Image captioning (IC), Image-Text Matching (ITM), Unrelatedness, Entity Extraction (EE), Visual Question Answering (VQA). We also asked GPT-4V to criticize its results (i.e., Self-critique).

Our work is inspired by recent works, such as \newcite{fu2023mme, ye2023mplugowl, liu2023mmbench} and more,  which leverage a small yet well-curated sample size of images to give instructions \cite{xu-etal-2023-multiinstruct, fu2023mme, yue2023mmmu, yu2023mmvet} to VLMs to perform VL tasks on those images. For example, \newcite{fu2023mme} assembled a set of 1,077 \textit{images} and corresponding \textit{instruction-answer pairs} to perform eleven perception and cognition tasks\footnote{These tasks are: coarse-grained object recognition (30, 60), fine-grained object recognition — recognizing movie posters (147, 294), celebrities (170, 340), scenes (200, 400), landmarks (200, 400), artworks (200, 400) —, optical character recognition (20, 40), commonsense reasoning (70, 140), numerical calculation (20, 40), text translation (20, 40), code reasoning (20, 40). Enclosed are \textit{\#images, \#instruction-answer pairs} for each sub-task.} in the MME benchmark. 

Notably, unlike \newcite{fu2023mme} who provide \textit{instruction-answer pairs} on top of the image at prompting time, we do not give any \textit{answers} to the VLM. By generating \textit{answers} for \textit{instructions}, we created new datasets for several VL tasks. In this study, we selected the tasks such that some of the tasks require the VLM to reason only about the \textit{image} while others require reasoning about both \textit{image and text} information (details in Section \ref{subsec:VisionAndLanguageTasks}). In addition, recognizing the lack of image-text datasets in Sw, Ur, we aimed at constructing datasets for many VL tasks.
\subsection{Data Source}
\label{sec:Data source}
\begin{figure}[!t]
\centering
\includegraphics[width=5.9cm]{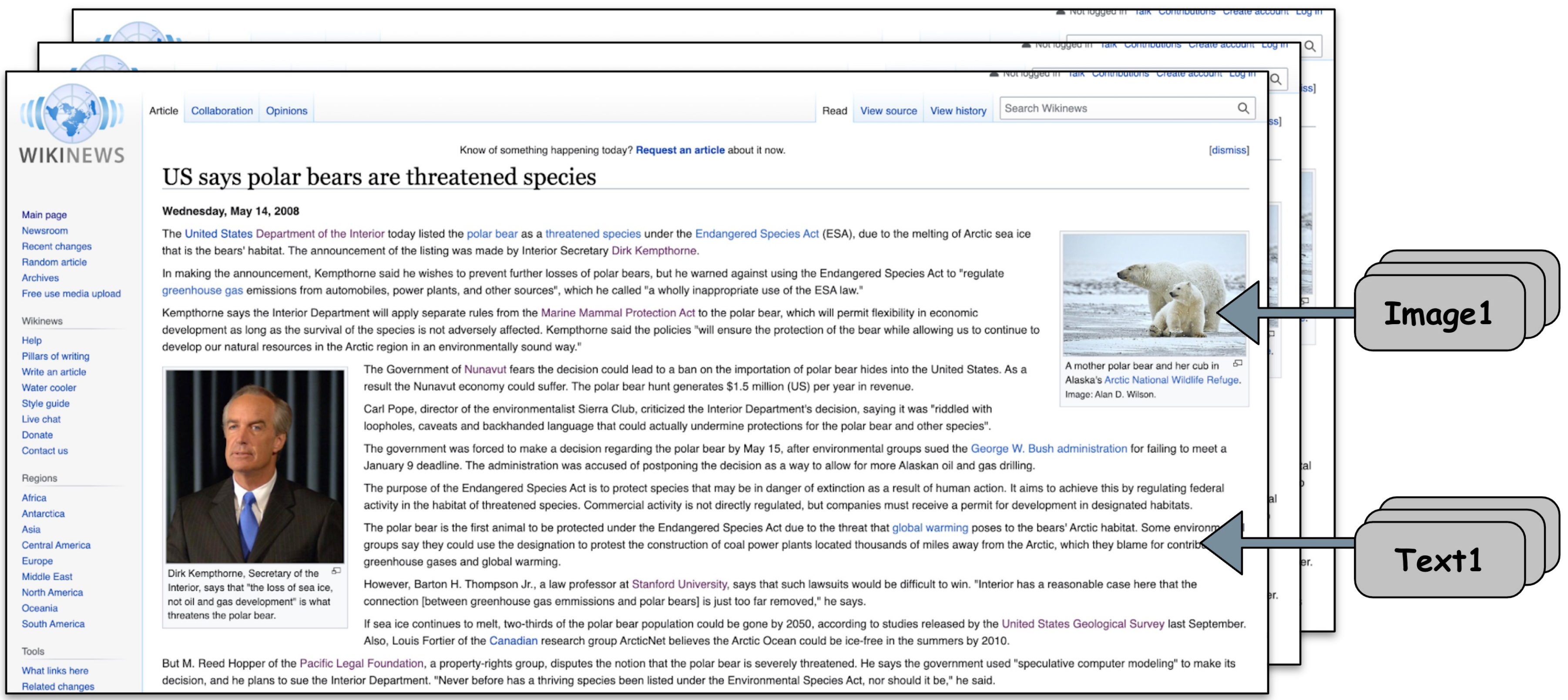} 
\caption{We gathered one image and all the text available in a \texttt{Wikinews} article. This is the definition of an \textit{image-text pair} in our study.}
\label{fig:DataSource}
\end{figure}

Our dataset consists of image-text pairs in En, Jp, Sw, and Ur. However, unlike the nature of image-text pairs found in image captioning \cite{lin2015microsoft, plummer2016flickr30k} datasets, in which a \textit{short} phrase called the \textit{caption} is assigned to an image, we seek to provide ``rich context'' about each image. Therefore, we gathered articles mainly from Wikinews\footnote{This is a link to Wikinews articles in English \url{https://en.wikinews.org/wiki/Main_Page}.}, and few articles from Wikipedia\footnote{Data dumps from Wikipedia are available at \url{https://dumps.wikimedia.org/enwiki/}.}. Specifically, we collected one image that was most relevant to the contents of the news article and the entire text inside the Wikinews article. This is our definition of an \textit{image-text pair}. Moreover, the article's text must be related to the image (see Figure \ref{fig:DataSource}). Under this setting, we provided a rich context for each image. 

After deciding on the data source, we needed to diversify our dataset. Hence, we gathered image-text pairs for ten categories: animals, products, buildings, locations, events, food, drinks, hobbies, works of art, and organization, because articles about these categories are readily available on Wikinews and Wikipedia. Our initial dataset includes 1,000 image-text pairs (100 per category), which we collected. Then, we randomly selected 20 images for each category; hence, the total is 200 image-text pairs used for VL analysis. Finally, we gathered data for 200 unique image-text pairs for each language across nine VL tasks.  
\begin{table*}[!t]
\footnotesize
\centering
\resizebox{\textwidth}{!}{\begin{tabular}{l}
\toprule
\textbf{Input Prompt in English} \\
\midrule
``[Put your text here]'' \\
Given the above text and image please answer the following questions one by one: \\
Q1. What objects are present in this image, and can you categorize them (e.g., furniture, electronic devices, etc.)? \\
Q2. Describe the overall scene depicted in this image. What is the setting, and what activities or events are taking place? \\
Q3. Identify any interactions or relationships between objects or entities in this image. How are they related or interacting? \\
Q4. Can you divide this image into different semantic regions and label each (e.g., sky, buildings, people, street)? \\
Q5. Provide a detailed, natural language description of what is happening in this image. \\
Q6. Given the content of this image, answer the following 4 sub-questions. \\
Q6.1  Extract the part of the text which closely matches the entities depicted in the image. \\
Q6.2 Which part of the text is not relevant to the image? \\
Q6.3 Mark all the spans of text which correspond to all the entities in the text above irrespective of the image provided. \\
Q6.4 What places are mentioned in the text or image above? Are they famous? If so, why? \\
Q7. On a scale of 1 to 5, how confident are you about your answer? \\

Consider one question at a time and give the answer. \\
Be as concise as possible while answering and avoid repetition. \\
For each question, please explain the rationale behind your answer. \\
Hence, use the following format when answering (Question, Answer, Rationale). \\
\bottomrule
\end{tabular}}
\caption{\label{Table:InputPrompt} \textit{English} prompt used to generate data for VL tasks.}
\end{table*}
\subsection{Vision-and-language Tasks}
\label{subsec:VisionAndLanguageTasks}
To study the visual and linguistic capacity of GPT-4V, we introduced nine tasks which either require only \textit{image}  information, that is, Task 1,2,3,4,5; or both \textit{image-and-text} information, that is, Task 6.1, 6.2, 6.3, 6.4 (We used this index, of Task 6.1, 6.2, 6.3, 6.4, instead of Task 6,7,8,9 because, at the time of creating the data, we needed to explicitly inform our evaluators that these four tasks require both \textit{image-and-text} information to evaluate the task correctly. Nevertheless, each task is stand-alone.). \\
\textbf{Task 1: Object Recognition (OR).} The object recognition or object detection task involves confirming the presence of an object \cite{lin2015microsoft}. In this study, we investigated the ability of GPT-4V to identify and categorize objects within an image based on their visual features. \textit{For example cat, bottle, car, etc.} (Illustration in Figure \ref{fig:TaskObjectRecognition} in Appendix.)\\
\textbf{Task 2: Scene Understanding (SU).} The task involves interpreting the context and the overall scene beyond just individual objects \cite{kafle2017analysis}. \textit{For example, A girl is sitting on a bench in a park.} (Illustration in Figure \ref{fig:SceneUnderstanding} in Appendix.)\\
\textbf{Task 3: Relationship Understanding (RU).} This task requires the model to identify, characterize, and reason about relationships between different objects within an image \cite{10.1007/s11263-016-0981-7}, which includes relationships like spatial proximity, interaction, ownership, causality, social, and more. \textit{For example, A girl sitting on a desk is feeding the cat, and there is a pet-owner relationship between the girl and the cat.} \\
\textbf{Task 4: Semantic Segmentation (SS).} This task involves dividing an image into parts with a semantic meaning \cite{lin2015microsoft}, such as identifying roads, buildings, and people in a street scene. \\
\textbf{Task 5: Image Captioning (IC).} In this task, the VLM is required to generate a natural language description of an image \cite{lin2015microsoft}, and such a description captures the scene's content, including objects, actions, relationships, emotions, and atmosphere. \textit{For example, This image contains a girl wearing a pink color skirt and feeding a white color cat.} \\
\textbf{Task 6.1: Image-Text Matching (ITM).} This task requires the VLM to comprehend which parts of the text correspond to the image \cite{xu-etal-2023-multiinstruct}. Given an image-text pair, we prompted the VLM to select the exact part of the text that best describes the image. \\
\textbf{Task 6.2: Unrelatedness (U).} This is a new task that we introduced. Herein, we prompted the VLM to select the exact part of the text that is \textit{not relevant} to the image when given an image-text pair (see Figure \ref{fig:Unrelatedness} in Appendix). \\
\textbf{Task 6.3: Entity Extraction (EE).} This task requires the VLM to extract all the entities/nouns in the text, given an image-text pair. \\
\textbf{Task 6.4: Visual Question Answering (VQA).} Under this task, the VLM needs to understand a natural language question about an image and generate appropriate answers \cite{agrawal2016vqa} by integrating visual understanding, language comprehension, and reasoning abilities. \\
\textbf{Self-critique (SC).} In addition to the above tasks, we prompted GPT-4V to rate the quality of its answers, resulting in \textit{self-critique} data.\\
\begin{figure*}[!t]
\centering
\includegraphics[width=0.50\linewidth]{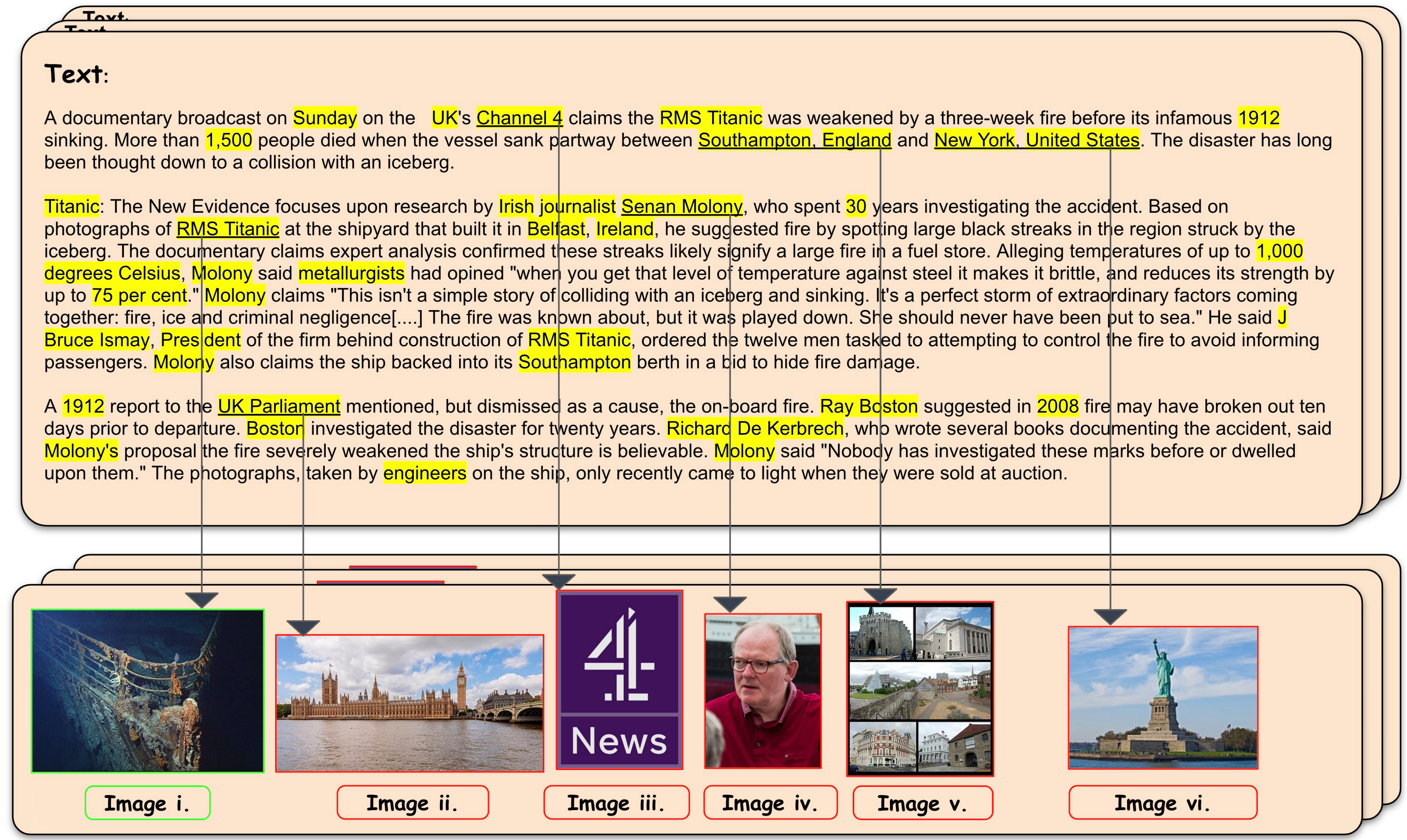} 
\caption{The original image from our data, \texttt{image i.} is shown on the left. The augmented images \texttt{image ii, image iii, image iv, image v}, and \texttt{image vi} are shown on the right. }
\label{fig:DataAugmentation}
\end{figure*}
\subsection{Prompting for Dataset Creation}
After deciding the categories and the tasks, we leveraged in-context learning and applied prompting to GPT-4V\footnote{We asked all the questions in one prompt because 1) the VLM's context window allows for generating the answer and rationale at once, 2)to save time and money of deploying GPT-4V.} \footnote{Our study is based on the GPT-4V version released in September 2023, and our dataset can be utilized for other VLMs in addition to GPT-4V.} to generate answers and rationales for each task, in four languages. We refined our prompts through a \textit{design-execute-output} cycle until satisfactory prompts that suited our study were obtained. An example prompt is shown in Table~\ref{Table:InputPrompt} (Prompt details in Figure \ref{fig:MultilingualPrompts3} in Appendix).
\subsection{Translation Tools}
In addition to employing native speakers\footnote{English is a widely spoken language globally and many people are familiar with its alphabet. This is not the case for the other languages in our study. Japanese has a distinct writing style (kanji, hiragana, katakana) and grammar. Swahili entails unique grammar rules, and it is a low-resource language. Hence, many translation tools do not support Swahili. Urdu has a distinct writing style: Nastaliq script and from right to left, different grammar rules,  and Urdu is low-resource too.} to translate the \textit{prompt} from En to Jp, Sw, Ur, we utilized readily available tools to translate the \textit{text} into each language. We employed DeepL\footnote{\url{https://www.deepl.com/translator}} for En-Jp, ChatGPT for En-Sw, and Google Translate\footnote{\url{https://www.google.com/search?client=safari&rls=en&q=google+translate&ie=UTF-8&oe=UTF-8}} for En-Ur translations. In each case, we chose the tool that native speakers agreed gave the best translations and representation of the native speakers' language. The quality of translations is shown in Table \ref{Table:TranslationAccuracy}.
\begin{table}[!t]
\footnotesize
\centering
\begin{tabular}{lllll}
\toprule
\textbf{Language} & En  & Jp  & Sw & Ur \\
\hline
\textbf{Accuracy (\%)}  & —  & 73.82  & 65.75 & 81.20 \\
\bottomrule
\end{tabular}
\caption{\label{Table:TranslationAccuracy} Accuracy of translations of \textit{texts} from En to Jp, Sw, and Ur, measured by native speakers.}
\end{table}
\begin{table}[!t]
\centering
\footnotesize
\begin{tabular}{ll}
\toprule
\textbf{\# Languages} &   4\\
\textbf{\# VL tasks per language} &   10*\\
\textbf{\# Questions per language} & 2,070 \\
\textbf{\# Image-text pairs per language} &   207\\
\textbf{\# Answer-rationale pairs per language} &   2,070\\
\textbf{\# Data samples per language} &   2,070\\
\midrule
\textbf{Total \#data samples}  & 7,210 \\
\bottomrule
\end{tabular}
\caption{\label{Table:DataStatistics} A summary of the characteristics of our new dataset. *The actual VL tasks are nine, but we also included the self-critique task. Only in Ur did we have 100 image-text pairs. (see full Table \ref{Table:FullDataStatistics} in Appendix \ref{Appendix: Full dataset stats}.)}
\end{table}
\paragraph{En to Jp, Sw, Ur Translations.} The average accuracy of translations for all \textit{texts} is shown in Table~\ref{Table:TranslationAccuracy}. Translations to Ur are the best, followed by Jp and Sw, respectively.
\subsection{Data Statistics}
\label{subsec:DataStatistics}
The statistics about visual-text data created in this study are shown in Table~\ref{Table:DataStatistics}. 
\subsection{Data Augmentation}
\label{sec:Data Augmentation}
Due to the ``rich text'' used in this study, it is the case that every \textit{text} contains lots of entities inside. Therefore, on top of one image which we assigned to each \textit{text}, we augmented the image-text pair by adding five new images, based on five entities randomly selected from the \textit{text}\footnote{The augmented image-text pairs will be used for further analysis especially the case where multiple \textit{images}  and a \textit{text} are presented to the VLM. Additionally, we will use the augmented data for multimodal NER analysis in the future.}(see Figure~\ref{fig:DataAugmentation}). 
\begin{figure*}[!t]
\centering
\includegraphics[width=0.60\textwidth]{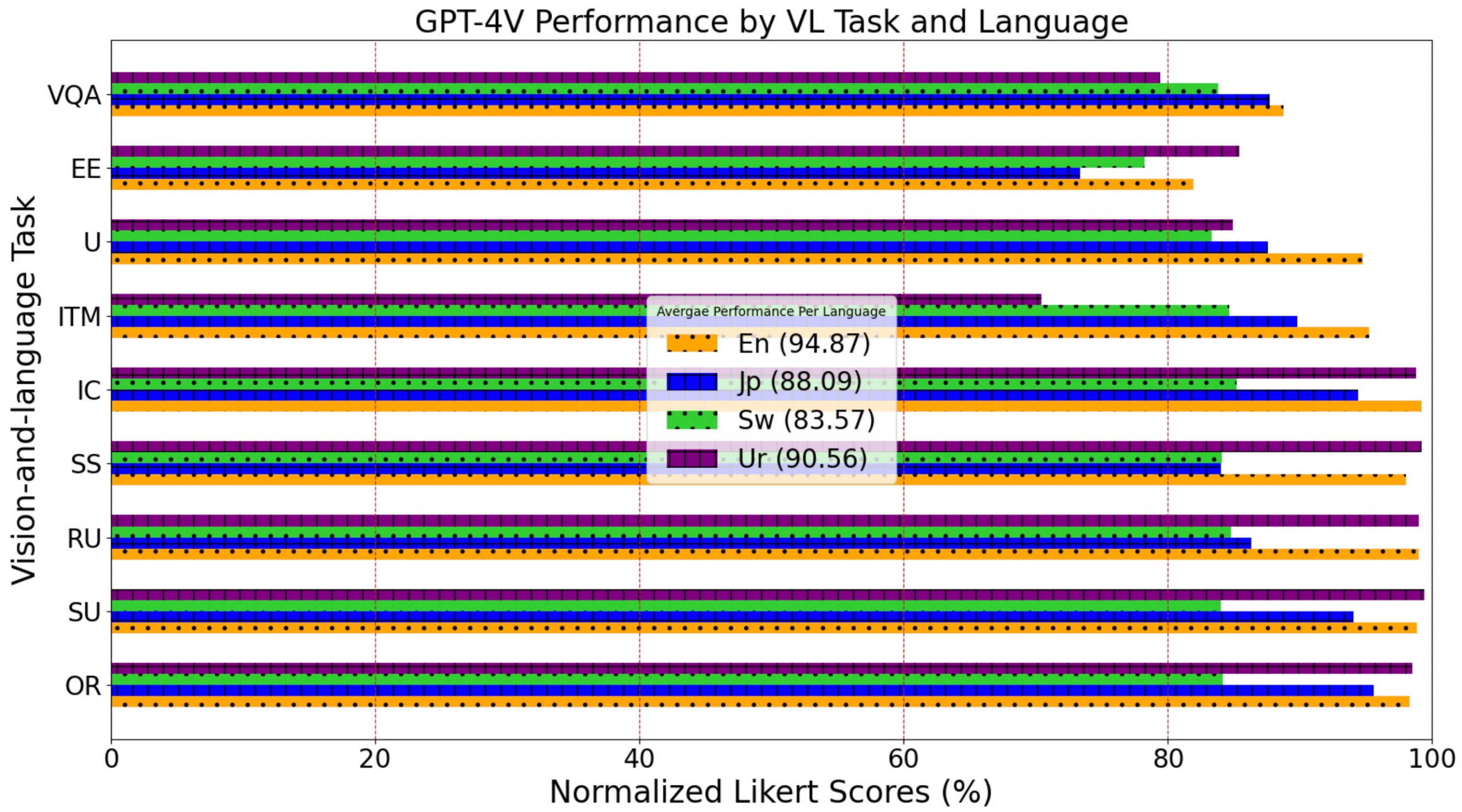} 
\caption{The performance of GPT-4V across all tasks as rated by native speakers of En, Jp, Sw, Ur. The scores are the normalized 5-scale Likert scores.}
\label{fig:MulitiLingualEvaluationPlot}
\end{figure*}
\section{Human Evaluation}
\label{subsec:HumanEvaluation}
\paragraph{Overview.} Due to the complex tasks investigated in this study, evaluating data samples is challenging. For example, our prompt in Table \ref{Table:InputPrompt} includes the question \textit{``Q1. What objects are present in this image, and can you categorize them (e.g., furniture, electronic devices, etc.)?''}, and the question \textit{``Q2. Describe the overall scene depicted in this image. What is the setting, and what activities or events are taking place?''} which look similar but require evaluators to think of ``object recognition'' and ``scene understanding'' tasks, respectively. Such situations necessitate that the evaluator has basic VL training and can make reasonable judgments based on the combination of image, text, and question information.
We recruited eight evaluators in total, two for each language. All the evaluators possess basic visual-linguistic knowledge and are instructed using well-documented evaluation guidelines. Two evaluators must be native speakers of En, Jp, Sw, and Ur to judge each data sample. 
Every evaluator spent an average of 12.5 hours during the evaluation process, and we compensated all evaluators based on the market price. 
\paragraph{Evaluation Guidelines.} The evaluation task comprised two main parts. In part 1, evaluators were asked to assign scores to the quality of En-Jp, En-Sw, and En-Ur translations of \textit{text}, on a Likert scale of 1 to 5\footnote{For clarity, 1 is the lowest score and 5 is the highest score.}. In part 2, we asked the evaluators to grade the quality of answers generated by GPT-4V for all \textit{questions} in the prompt\footnote{Again, we utilized a Likert scale with scores from 1 to 5 in which 1 is the lowest score, and 5 is the highest score.}. Exact details about evaluation guidelines are shown in Appendix \ref{Appendix:EvaluationGuidelines}.
\begin{table*}[!t]
\tiny
\centering
\resizebox{\textwidth}{!}{\begin{tabular}{lllllllllll}
\toprule
\textbf{Category} & Anm. & Pdt. & Bdg. & Loc. & Evt. & Art. & Org. & Drk. & Hob. & Fd.\\
\midrule
\textbf{Partial match}   & 0.2644  & 0.2059 & 0.1543 & 0.2474 & 0.2225 & 0.12105 & 0.1362 & 0.1006 & 0.1195 & 0.2488 \\
\bottomrule
\end{tabular}}
\caption{\label{Table:RelationshipUnderstandingPM} Partial match scores between GPT-4V and LLaVA-1.5 per category (Animals, Products, Buildings, Locations, Events, Work of Art, Organisations, Drinks, Hobbies, Food, respectively). }
\end{table*}
\section{Multilingual Evaluation} 
\label{subsec: Multilingual Evaluation}
A multilingual evaluation for each task, measured by En, Jp, Sw, and Ur native speakers, is shown in Figure~\ref{fig:MulitiLingualEvaluationPlot}. GPT-4V did exceptionally well in En, and its performance in Jp and Ur was comparable. GPT-4V struggled more with Sw. \\
\textbf{En.} Generally, GPT-4V flourished across all tasks with an average accuracy of 94.81\%. Notably, for tasks in which reasoning only about the \textit{image} is needed,  OR, SU, RU, SS, IC, GPT-4V performed very well. There was a slight dip in performance when GPT-4V reasoned over combined image-and-text information under ITM, U, EE, and VQA tasks. GPT-4V performed best at OR (99.20\%) and slightly struggled only with EE (81.95\%). \\
\textbf{Jp.} The average accuracy of GPT-4V is 88.09\%. The best-performing task was OR (95.56\%), while the worst was EE (73.33\%). A similar phenomenon was observed in which reasoning over images was easier than reasoning over images and text. \\
\textbf{Sw.} The average performance of GPT-4V in \textit{Sw} was lower than in \textit{En}. The average accuracy was \textit{83.57\%}. IC (85.22\%) was the best-performing task, yet EE (78.21\%) was the worst. \\
\textbf{Ur.} The performance of GPT-4V in \textit{Ur} was lower than in \textit{En}, and the average accuracy was 90.56\%. Similar to English, tasks that required reasoning over only the image were easy for GPT-4V. Yet, GPT-4V had a slightly lower performance for tasks that necessitated reasoning over both the image and text. \\
\section{Automatic Evaluation}
\label{sec:Methods}
To investigate how close to GPT-4V the responses from open source models are, we fine-tuned LLaVA 1.5  on the test set\footnote{We used the test set because it contains high-quality human-evaluated data samples, sufficient for usage as the ground truth when comparing with other VLMs' responses.} of our data containing 100 image-text pairs on the task of \textit{relationship understanding}\footnote{RU task was picked at random, out of all tasks.}. We compared the responses of GPT-4V (a closed-source VLM) to responses of LLaVA 1.5 (an open-source VLM) via \textit{partial match}.
\textbf{GPT-4V} is considered to be the most powerful VLM. %
The \textbf{LLaVA 1.5}\footnote{The official release for all versions of LlaVa is available at \url{https://github.com/haotian-liu/LLaVA/blob/main/docs/MODEL_ZOO.md\#llava-v16}.} variant used in our study is the 4-bit quantized version of \textit{LLaVA-1.5-13B} \cite{liu2023improvedllava}.
Details in Appendix \ref{Appendix:LLaVANeXTmPLUG-Owl2}.
\section{Results}
\label{sec:Results}
\paragraph{Relationship Understanding.}
\label{subsec:RelationshipUnderstanding}
When fine-tuned on our dataset, LLaVA 1.5 partial match scores with GPT-4V are shown in Table \ref{Table:RelationshipUnderstandingPM}. The average score across all categories is 0.1821, indicating little similarity between GPT-4V and LLaVA 1.5 responses,  when the same image-text pairs are provided to both models\footnote{This is insightful for further analysis.}.
\section{Discussion}
\label{sec:Discussion}
\paragraph{General Observations.} On top of scores reported in Section~\ref{sec:Results}, we asked the following questions to the evaluators: \textit{(i) How coherent are the answers provided by GPT-4V? (ii) Would you (as a native speaker) give such answers as GPT-4V? If yes/no, please justify. (iii) Which task do you think GPT-4V performed best? (iv) In which task did GPT-4V perform worst? (v) What areas of improvement did you observe? }

These are the observations made by the native speakers.
\textbf{(i) How coherent are the answers provided by GPT-4V?}
Answer: ``The responses given by GPT-4V show a significant variation in coherence depending on the type of input it deals with. GPT-4V's responses were generally coherent and on point when the questions were based purely on images. However, the performance dipped noticeably when the task required interpreting both text and images together. The answers in such cases tended to include irrelevant or redundant information, which detracted from their overall relevance and coherence.''
\textbf{(ii) Would you (as a native speaker) give such answers as GPT-4V? If yes/no, please justify}
Answer: ``Personally, I wouldn't give answers like GPT-4V does, especially regarding the text+image tasks. The responses often felt artificial and lacked the natural flow or intuition a human might bring to interpret combined inputs. There's a distinct pattern to GPT-4V's responses that makes them recognizable as machine-generated, lacking the variability and adaptiveness you'd expect from a human''.
\textbf{(iii) In which task do you think GPT-4V performed best?}
Answer: ``GPT-4V shines in purely visual tasks like image captioning, scene understanding, relationship understanding, etc. It seems to have a better grip on interpreting images alone, where its responses are more focused and aligned with the queries. This is where it performed best, showing a good understanding of visual content without the complication of integrating textual information. For Example: For img \#6 GPT-4V performs well for the task that just includes Q1 to Q5 but from Q6.1 to Q6.4 the performance declined.''
\textbf{(iv) In which task did GPT-4V perform worst?} Answer: ``On the flip side, entity extraction tasks' worst performance was observed, especially when they involved both text and images. GPT-4V struggled to identify and highlight the relevant entities accurately. Instead of pinpointing specific items or figures, it often provided broad, generalized answers or extracted whole sentences, missing the mark on specificity and accuracy, particularly evident in tasks where precise identification of entities from images and accompanying text was crucial. We can see an example of this shortfall in image \#19, where GPT-4V's responses were more about citing sentences than identifying discrete entities, indicating a significant area for improvement in its ability to process and integrate multi-modal information effectively.''
\textbf{(v) What areas of improvement did you observe?}
From our observations, there are a few key areas where GPT-4V could use some improvements.
        \textbf{ (a) Handling combined Text and Image Inputs:} GPT-4V didn't perform as well when it had to deal with both text and images together. The answers it provided in these cases were often filled with irrelevant details. It would be beneficial if GPT-4V could better integrate information from both sources to provide more coherent responses.
        \textbf{ (b) Naturalness and Variability of Responses:} The responses from GPT-4V sometimes felt too mechanical and predictable, lacking the variability and intuitive grasp that a human response would have. Making the model's output more varied and human-like would greatly improve its interactions, making it harder to distinguish from a human's response.
        \textbf{ (c) Accuracy in Entity Extraction:} In this task, especially with text and images, GPT-4V often missed the mark. It either provided too broad answers or extracted entire sentences instead of focusing on specific entities. A more precise approach to identifying and extracting entities from complex inputs would enhance its utility in this task.
        \textbf{(d) Reducing Repetitiveness:} The patterned nature of GPT-4V's responses was a bit of a giveaway that you're talking to a machine. Working on making its language generation more dynamic and less prone to falling into repetitive patterns would make the conversation flow more naturally, akin to human interaction.

Focusing on these areas would not just make GPT-4V's answers more coherent and relevant, especially in mixed media tasks, but also enhance its overall interaction quality, making it an even more powerful tool for a wide range of applications.
\paragraph{Relevance of Rationales.} In this study, all evaluators agree that rationales provided more evidence necessary for the evaluators to make an informed decision during evaluation. 
\paragraph{Impact of Jp, Sw, Ur Translation Quality.} We hypothesize that the quality of translations impacted the ability to accurately comprehend \textit{text} fed into GPT-4V. 
As indicated in Section~\ref{sec:Results}, \textit{Ur} translations were the best, followed by \textit{Jp} and \textit{Sw}, respectively. This is probably one of the reasons why we can see that the average GPT-4V performance was best in \textit{Ur} followed by \textit{Jp} and \textit{Sw}.
\paragraph{ GPT-4V Self Critique.} On a Likert scale of 1 to 5, the average score of \textit{self-confidence} for generating \textit{answer, rationale} pairs for all VL tasks in English was 4.826. This indicates that GPT-4V is over-confident.  
\section{Conclusion}
\label{sec:Conclusion}
In response to the growing need to measure the VL strengths of VLMs, we constructed multilingual datasets containing image-text pairs for En, Jp, Sw, and Ur for nine VL tasks, including a new task \textit{unrelatedness}. By collecting 1,000 image-text pairs in total for ten categories, which will be publicly available, we further selected 200 samples (20 per category) for the construction of datasets through prompting GPT-4V. In addition, we augmented each image-text pair with five more images. 

By utilizing the 200 data samples, human evaluators rated the quality of our constructed datasets to establish their suitability for VL tasks. The human evaluators' feedback further highlighted VLMs' strengths and weaknesses and a performance gap across En, Jp, Sw, and Ur. Based on this feedback, we constructed a high-quality test set that we can use for automatic evaluation across nine tasks. To demonstrate the effectiveness of our dataset, we chose the \textit{relationship understanding} task. We fine-tuned LLaVA 1.5 on a subset of our dataset and measured the partial match between the GPT-4V and LLaVA 1.5. \textit{We will publicly release all datasets and evaluations after publication.} 
\section*{Limitations}
\label{sec:Limitations}
\paragraph{Evaluation Criteria.} Throughout this study, our evaluation is based on a sample of 200 image-text pairs out of the 1,000 pairs we collected. The samples are evenly distributed across all 20 pairs. We would like to utilize all the data for further analysis. Moreover, automatic evaluation reported in this work covers one VL task (i.e., relationship understanding), utilizing a subset of the entire dataset.
\paragraph{Human Evaluation Cost.} Embarking on such a project comes with a hefty price because recruiting human evaluators is expensive. 
\paragraph{No GPT-4V Details.} There needs to be details about the architecture of GPT-4V, which leaves us in the dark about what aspects of this model contribute to its impeccable performance.

\bibliography{main}
\bibliographystyle{acl_natbib}

\newpage
\appendix

\section{Appendix}
\label{sec:appendix}

\subsection{Set-of-Mark Prompting} 
\label{Appendix:setofmark}
Set-of-Mark (SoM) is a visual prompting technique introduced by \newcite{yang2023setofmark} to improve the grounding abilities of VLMs. By assigning ``marks'' to different objects inside the image in the form of alphanumeric labels, masks, or boxes, the VLM is more capable of correctly answering any questions about the image. We utilized this technique in the early stages of our study. 
Though SoM\footnote{This site \url{https://huggingface.co/spaces/Roboflow/SoM} provides a platform to augment images with different marks, for example, numbers, letters, bounding boxes, and masks.} improves the visual grounding ability of VLMs; SoM did not offer significant gains in the quality of answers generated by GPT-4V during the preliminary stage of our study. Hence, we did not extensively deploy SoM in this study, and we leave that for future work.
\subsection{LLaVA 1.5}
\label{Appendix:LLaVANeXTmPLUG-Owl2}
In our study LLaVA 1.5 \cite{liu2023improvedllava} is used, and here are the details. LlaVa 1.5 consists of three main components, that is, a vision encoder, MLP connector, and a pre-trained language model. Like other VLMs, this model was trained in two stages, the pre-training stage in which visual features contained inside the image were aligned with the word embedding space of the language model, followed by the visual instruction tuning stage in which visual instructions were leveraged to tune the model so that it can comprehend user instructions about images. Notably, the language encoder is  Vicuna-13B-v1.5, while the vision encoder is CLIP-L-336px.
Due to the computational cost involved in fine-tuning, we applied 4-bit quantization to LLaVA 1.5.
\subsection{Evaluation Guidelines}
\label{Appendix:EvaluationGuidelines}
The evaluation task comprised two main parts. In part 1, evaluators were asked to assign scores to the quality of En-Jp, En-Sw, and En-Ur translations of \textit{text}, on a Likert scale of 1 to 5\footnote{For clarity, 1 is the lowest score and 5 is the highest score.}. In part 2, we asked the evaluators to grade the quality of answers generated by GPT-4V for all questions Q1 to Q6 in the prompt\footnote{Again, we utilized a Likert scale with scores from 1 to 5 in which 1 is the lowest score, and 5 is the highest score.}.

Precisely, here are the evaluation guidelines in part 1:
\begin{itemize}
    \item Please refer to the English text and rate the quality of the translated text from 1 to 5.
    \item Note that the 1 is the lowest score and 5 is the highest score
\end{itemize}

In addition, these are the evaluation guidelines which we passed on to native speakers in part 2:
\begin{itemize}
    \item Taking into consideration, the combination of \textbf{image, text, question, answer} and \textbf{rationale} generated by GPT-4V for each question Q1 to Q6, please do the following:
    \item Read carefully the \textbf{answer, rationale} pair and make a reasonable judgement if the answer best represents the information provided
    \item For each question, these are the guidelines;
    \begin{itemize}
        \item Q1. Has GPT-4V identified and categorized all the objects in image?
        \item Q2. Has GPT-4V interpreted the context and the overall scene, beyond just individual objects?
        \item Q3. Has GPT-4V accurately recognized the relationships and interactions between different objects in a scene? 
        \item Q4. Has GPT-4V appropriately divided the image into parts with a semantic meaning?
        \item Q5. When generating a natural language description of the image, did GPT-4V cover all the objects in the image?
        \item Q6.1 Did GPT-4V extract accurately the part of the text which closely matches the entities shown in the image? 
        \item Q6.2 Looking at the image and the text, did GPT-4V correctly report the text which is not relevant to the image? 
        \item Q6.3 Ignoring the image, did GPT-4V mark all the spans of text which correspond to the entities in the text provided? 
        \item Q6.4 Are the places reported by GPT-4V actually famous and are they present in the text or image provided?
    \end{itemize}
\end{itemize}
\subsection{Benchmark Details}
\label{Appendix:benchmarkdetails}
MMBench \cite{liu2023mmbench} measures twenty VLM abilities across both perception and reasoning problems. The specific perception tasks include spatial relationship, attribute comparison, action recognition, attribute recognition, object localization, celebrity recognition, OCR, image style, image scene, image recognition, image quality, and image topic. Similarly, reasoning abilities include future prediction, structuralized image-text understanding, physical property, function reasoning, identity reasoning, social relation, physical relation, and natural relation.  

SEED-Bench \cite{li2023seedbench} measures twelve VLM abilities, namely action prediction, action recognition, procedure understanding, scene understanding, instance identity, instance attribute, instance location, instance counting, spatial relation, instance interaction, visual reasoning, and OCR. 

MM-Vet \cite{yu2023mmvet} evaluates the performance of VLMs on combinations of various abilities related to knowledge, language generation, math, OCR, recognition, and spatial awareness.
\subsection{Full dataset statistics}
\label{Appendix: Full dataset stats}
Shown in Table \ref{Table:FullDataStatistics}.
\begin{table*}[!t]
\centering
\footnotesize
\begin{tabular}{llllll}
\toprule
& \textbf{Language} &   {\bf English}   & {\bf Japanese}   & {\bf Swahili} & {\bf Urdu} \\
 & & & & & \\
\textbf{Category} & \textbf{Task} &  &  & &  \\
\midrule
Image               & OR           & 207 & 207 & 207 & 100 \\
                    & SU          & 207 & 207 & 207 & 100 \\
                    & RU   & 207 & 207 & 207 & 100 \\
                    & SS        & 207 & 207 & 207 & 100 \\
                    & IC             & 207 & 207 & 207 & 100 \\
\midrule
                    & EE           & 207 & 207 & 207 & 100 \\
Image-and-text & ITM         & 207 & 207 & 207 & 100 \\
                    & U                    & 207 & 207 & 207 & 100 \\
                    & VQA   & 207 & 207 & 207 & 100 \\
\midrule
    &           Self-critique   & 207 & 207 & 207 & 100 \\
\midrule
\textbf{Total}    &  & 2,070 & 2,070 & 2,070 & 1,000 \\
\bottomrule
\end{tabular}
\caption{\label{Table:FullDataStatistics}
The number of data samples created per task, in each language. We categorize the tasks based on the \textit{type} of information required to actually accomplish the task. Each data sample is an \textit{Answer, Rationale} combination.} 
\end{table*}
\subsection{Multilingual Prompts}
\label{Appendix:GPT-4InputPromptOnlyText}
\begin{figure*}[!t]
\centering
\includegraphics[width=0.95\linewidth]{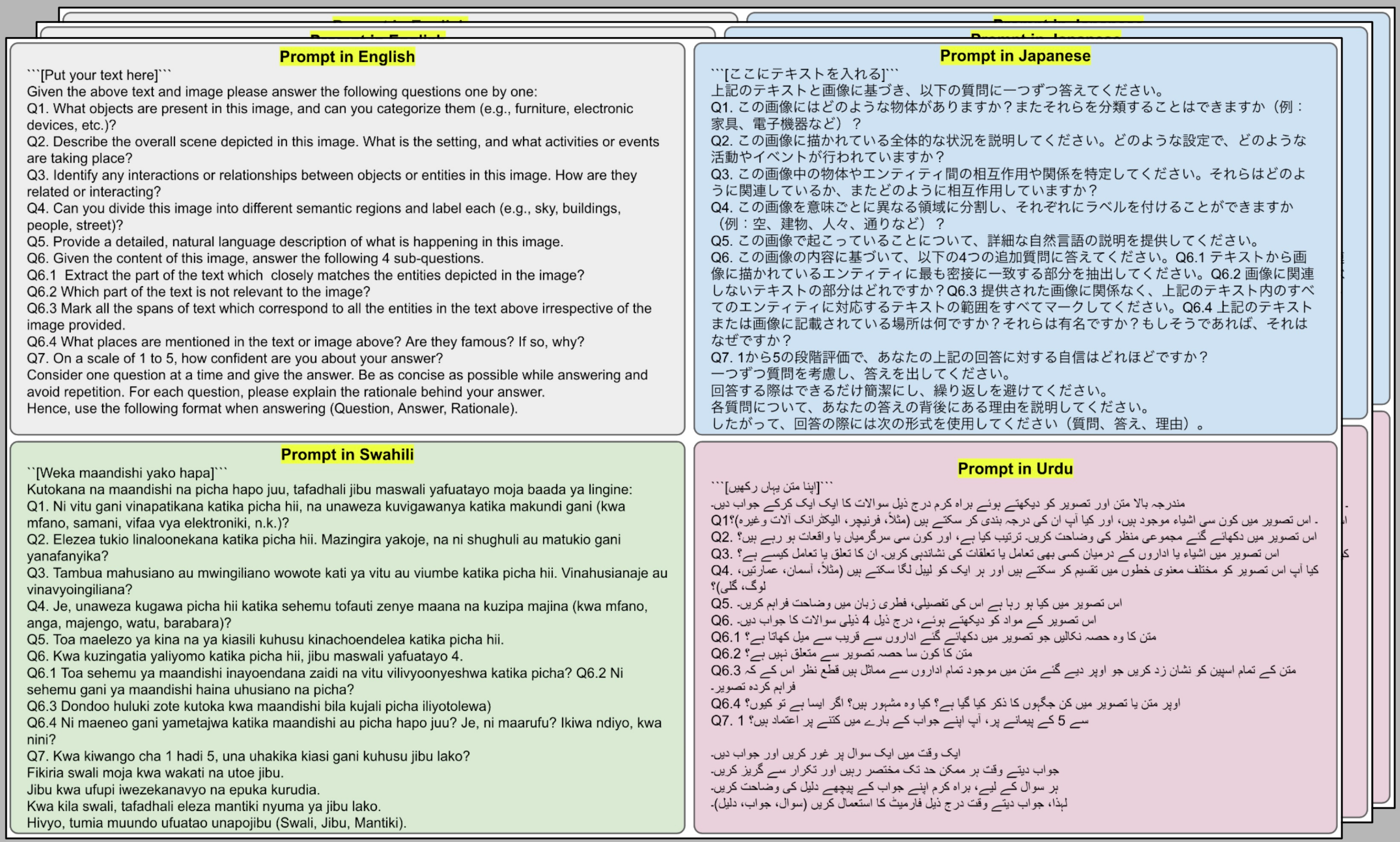} 
\caption{ The prompts which we used to construct data in each of the four languages: En, Jp, Sw, and Ur.} 
\label{fig:MultilingualPrompts3}
\end{figure*}

Our multilingual prompts used to gather data for each VL task are shown in Figure~\ref{fig:MultilingualPrompts3}.
\subsection{Input Prompt}
\label{Appendix:IntroFigureExpanded}
The complete version of Figure \ref{fig:IntroFigure} in Section \ref{sec:Introduction} is shown in detail in Figure~\ref{fig:IntroFigureExpanded} below.
\begin{figure*}[!t]
\centering
\includegraphics[width=0.95\linewidth, height=9cm]{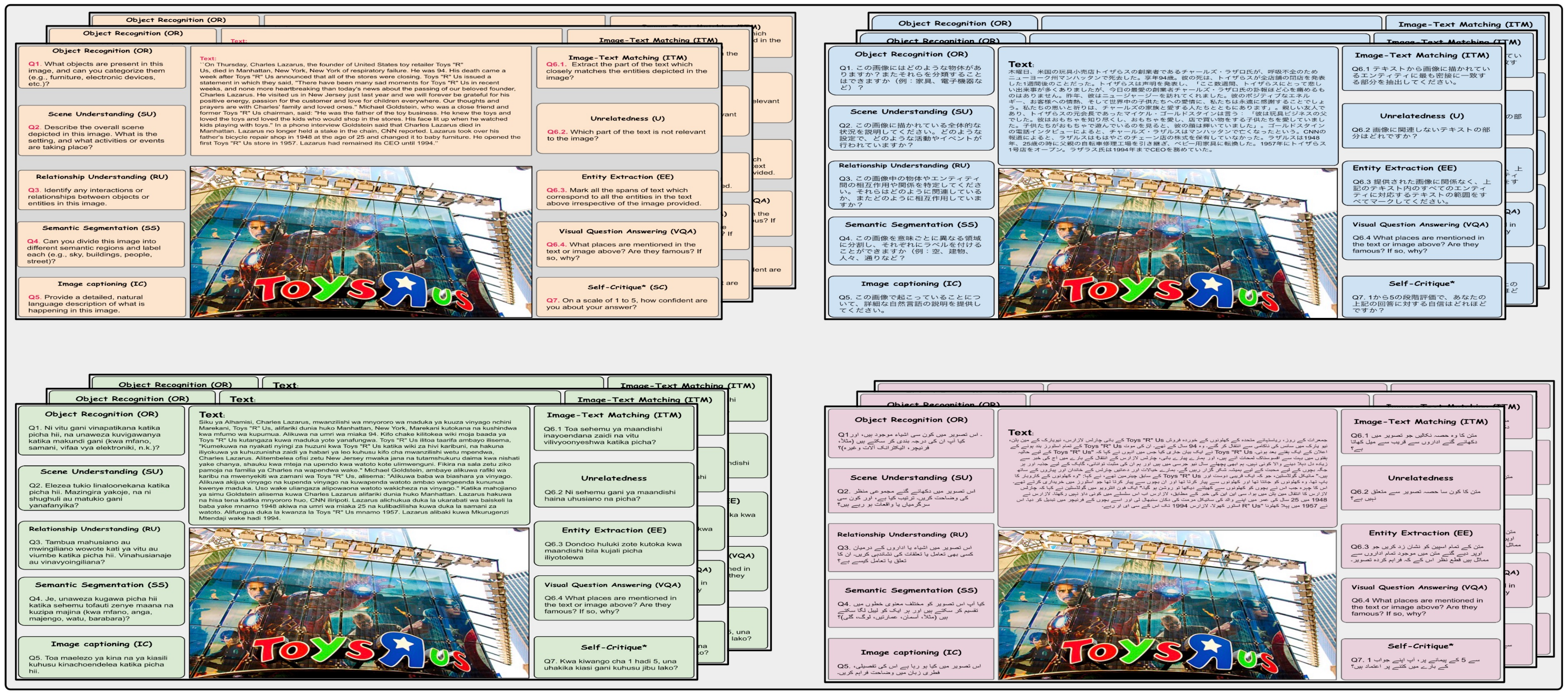} 
\caption{The expanded version of Figure~\ref{fig:IntroFigure}. Clockwise from top left: \textbf{English (En), Japanese (Jp), Urdu (Ur), Swahili (Sw)} prompts. The input to GPT-4V includes \textbf{text, image, questions Q1-Q7}.   }
\label{fig:IntroFigureExpanded}
\end{figure*}
\subsection{Self Critique Criteria}
\label{SelfCritiqueCriteria}
This is the criteria utilized by GPT-4V to calculate its confidence score, which defines the \textit{self-critique} in this study.
\begin{itemize}
    \item\textbf{Accuracy.} The extent to which the information provided is correct and factual.
    \item Relevance. How closely the answers pertain to the questions asked, including the presence of information in the text or image that directly supports the answer.
    \item \textbf{Completeness.} Whether the answer comprehensively addresses all parts of the question without omitting key details.
    \item \textbf{Clarity.} The degree to which the answer is understandable and unambiguous.
    \item \textbf{Evidence Support.} The amount of evidence available from the provided text or image that can be used to substantiate the answer.
\end{itemize}
\subsection{Swahili, Urdu Speakers}
\label{SwahiliUrduSpeakers}
As of this writing, the number of Swahili speakers is estimated at more than  200 million\footnote{Wikipedia page for Swahili \url{https://en.wikipedia.org/wiki/Swahili_language}.}, while the number of Urdu speakers stands at about 230 million\footnote{Wikipedia page for Urdu \url{https://en.wikipedia.org/wiki/Urdu}.} speakers.

\subsection{Unrelatedness Task}
\label{Appendix:Unrelatedness}
\begin{figure*}[!t]
\centering
\includegraphics[width=0.95\linewidth]{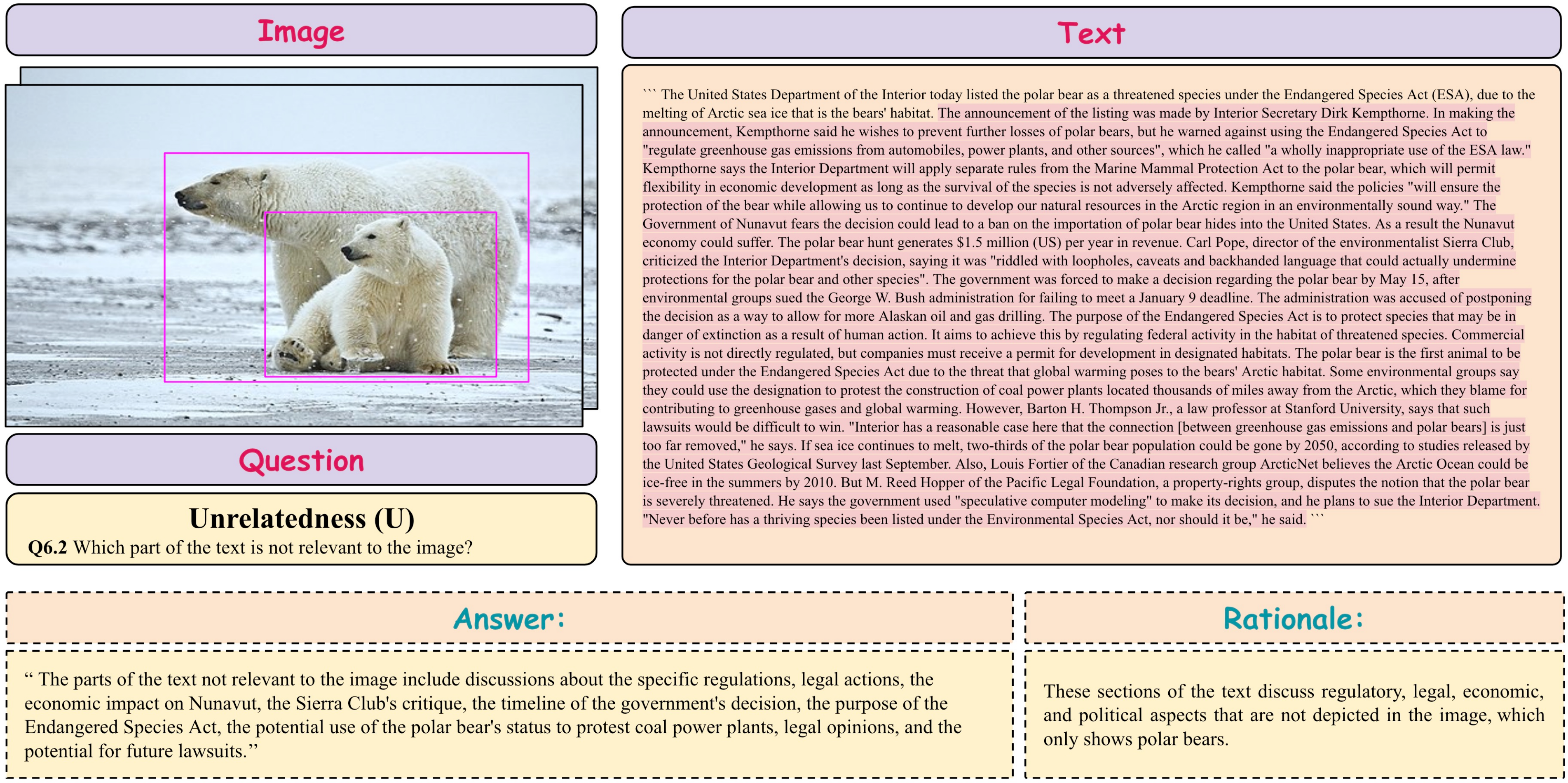} 
\caption{The new VL task introduced in this work is \emph{Unrelatedness}. Most of the \textit{Text} in this prompt is unrelated to the \textit{Image}.}
\label{fig:Unrelatedness}
\end{figure*}

The new VL task introduced in this work is \textit{Unrelatedness} and it is shown in Figure~\ref{fig:Unrelatedness}.
\subsection{Vision-and-language Tasks}
\label{Appendix: DataCreationPerLanguage}
\begin{figure*}[!t]
\centering
\includegraphics[width=0.95\linewidth]{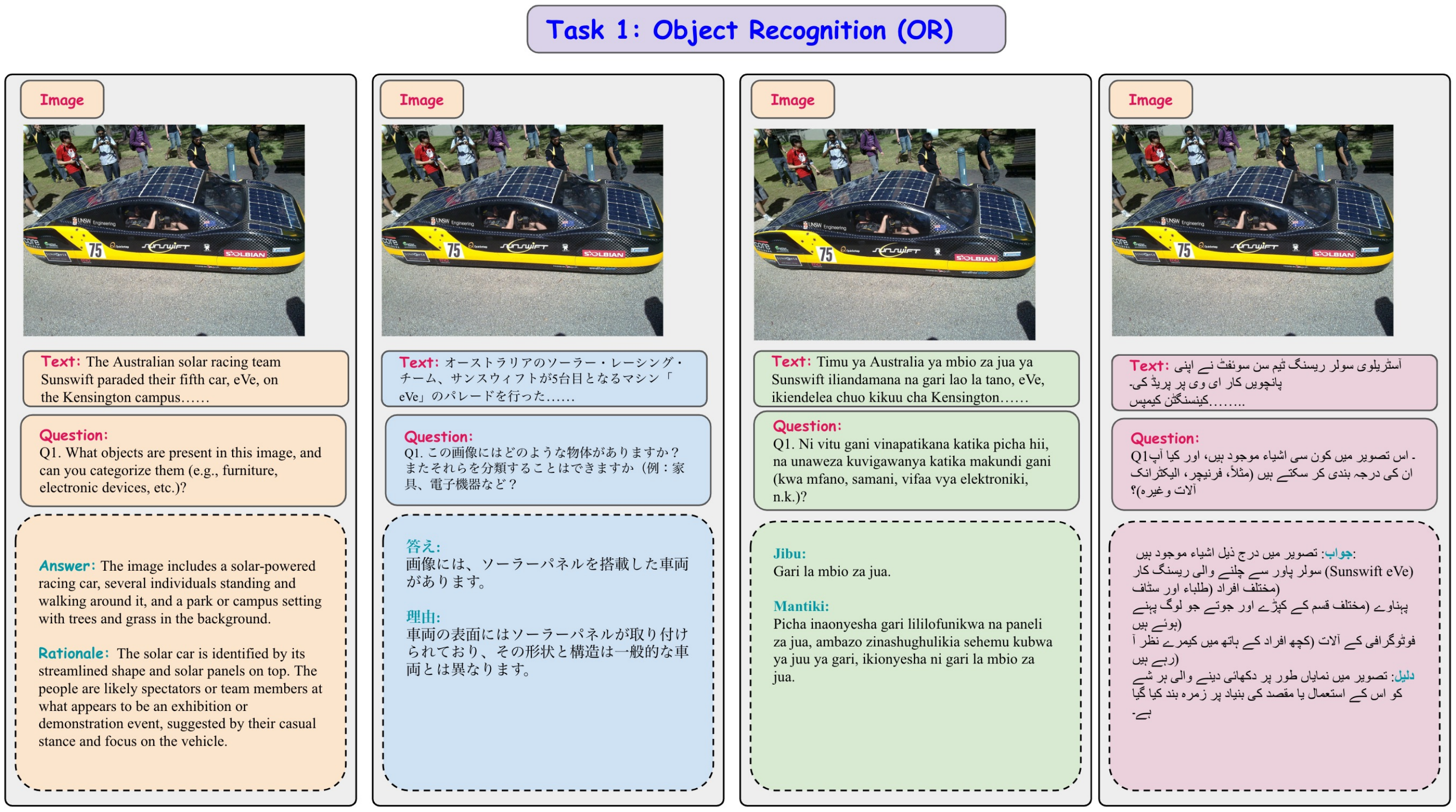} 
\caption{Example of an object recognition sample in En, Jp, Sw, Ur in our dataset.}
\label{fig:TaskObjectRecognition}
\end{figure*}

\begin{figure*}[!t]
\centering
\includegraphics[width=0.95\linewidth]{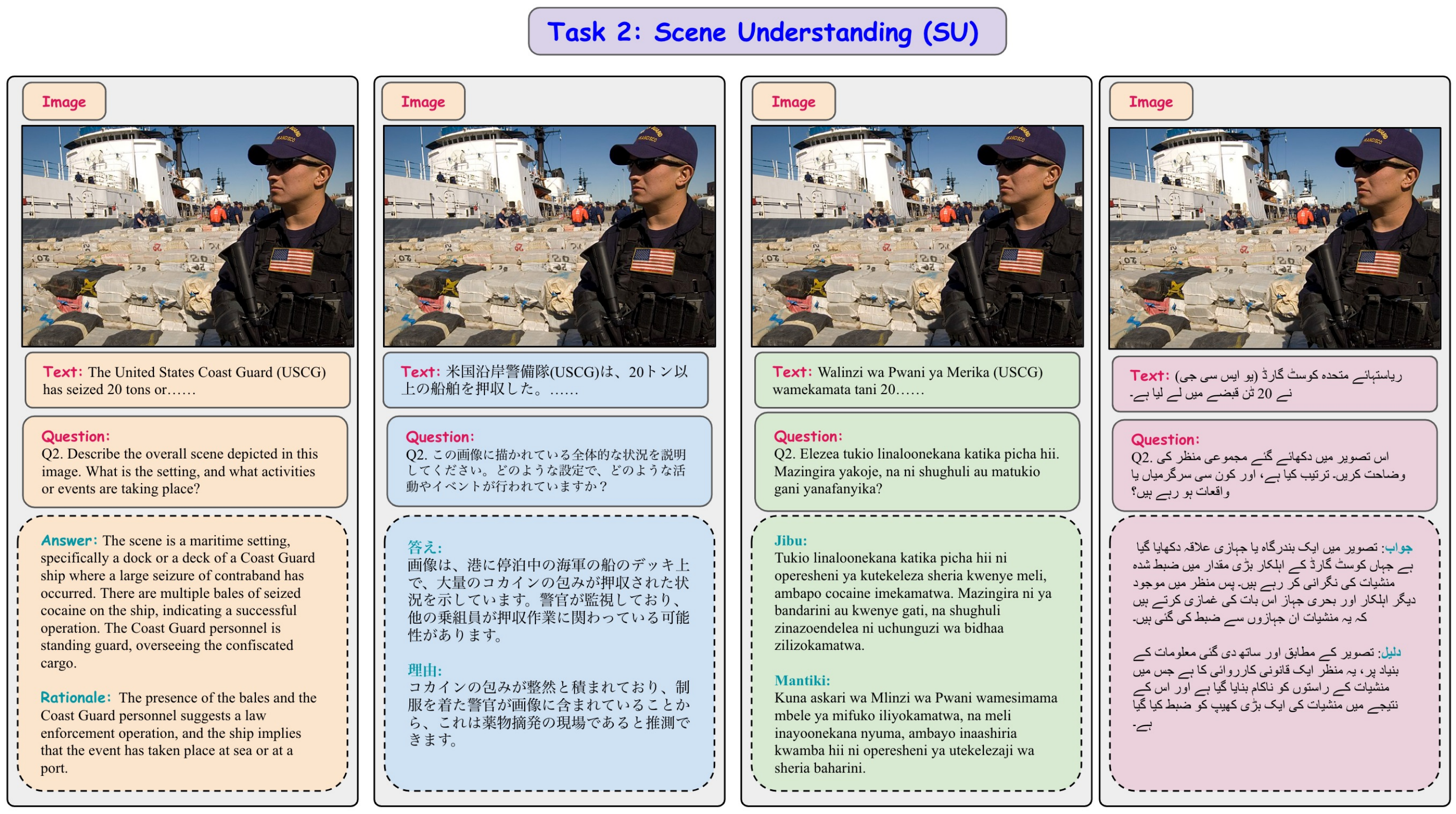} 
\caption{Example of a scene understanding sample in En, Jp, Sw, Ur in our dataset.}
\label{fig:SceneUnderstanding}
\end{figure*}

In this section, we illustrate the tasks introduced in our study. Moreover, owing to space limitations, we have shown only two tasks: \textit{object recognition} in Figure \ref{fig:TaskObjectRecognition} and \textit{scene understanding} in Figure \ref{fig:SceneUnderstanding}.

\end{document}